\documentclass[nohyperref]{article}

\usepackage{microtype}
\usepackage{graphicx}
\usepackage{subfigure}
\usepackage{booktabs} 

\usepackage{hyperref}

\usepackage[accepted]{icml2022}

\usepackage{amsmath}
\usepackage{amssymb}
\usepackage{mathtools}
\usepackage{amsthm}
\usepackage{wrapfig}
\usepackage[capitalize,noabbrev]{cleveref}

\theoremstyle{plain}

\theoremstyle{definition}

\theoremstyle{remark}

\usepackage[textsize=tiny]{todonotes}

\icmltitlerunning{Safe RL with Probabilistic CBFs}

\begin{document}

\onecolumn
\icmltitle{Safe Reinforcement Learning with Probabilistic\\ Control Barrier Functions for Ramp Merging}



\icmlsetsymbol{equal}{*}

\begin{icmlauthorlist}
\icmlauthor{Soumith Udatha}{equal,yyy}
\icmlauthor{Yiwei Lyu }{yyy}
\icmlauthor{John Dolan}{yyy}
\end{icmlauthorlist}

\icmlaffiliation{yyy}{Carnegie Mellon University, Pittsburgh, United States}

\icmlcorrespondingauthor{Soumith Udatha}{sudatha@andrew.cmu.edu}

\icmlkeywords{Safe Reinforcement Learning}

\vskip 0.3in



\printAffiliationsAndNotice{}  

\begin{abstract}
Prior work has looked at applying reinforcement learning and imitation learning approaches to autonomous driving scenarios, but either the safety or the efficiency of the algorithm is compromised.With the use of control barrier functions embedded into the reinforcement learning policy, we arrive at safe policies to optimize the performance of the autonomous driving vehicle. However, control barrier functions need a good approximation of the model of the car. We use probabilistic control barrier functions as an estimate of the model uncertainty. The algorithm is implemented as an online version in the CARLA \cite{Carla_Sim} Simulator and as an offline version on a dataset extracted from the NGSIM Database. The proposed algorithm is not just a safe ramp merging algorithm, but a safe autonomous driving algorithm applied to address ramp merging on highways.
\end{abstract}

\section{Introduction}
\label{submission}

It is a well-known fact that real-word dynamics are more complicated than can be easily modeled in an analytical form
in  simulators \cite{driver-behavior}. This disparity between the real dynamics and
the approximations involved becomes more prominent when
dealing with safety-critical systems. To determine safety of a
system we need to have a good idea of the world dynamics
involved.\\
The interaction of self-driving cars with other human
drivers can be treated as a safety-critical system.\cite{yiwei2021}. In the
area of autonomous vehicle control and planning (\citeauthor{Gu2015}, \citeauthor{7040372}, \citeauthor{luo2016distributed}),
safety is always the primary focus. 
Ramp merging is a typical, relatively simple scenario where
autonomous vehicles interact with humans. Even for a simple driving scenario,
freeway on-ramp merging areas are high-risk areas for motor vehicle crashes and conflicts due to the variety of driving styles \cite{9557770}.
Human drivers introduce uncertainty into merge scenarios, and Autonomous Vehicles (AV) will collide if they can’t plan and be controlled safely \cite{Zhu2021-vd}. It is therefore important that we address model uncertainty while modeling the control and planning of self-driving cars.\\
In situations with rule-based systems designed to adhere to various driver behaviors on a highway, strictly following the rules to maintain safety could result in unsafe situations, where the autonomous vehicle might brake for an instant when going very close
to a vehicle, which may not be anticipated by the human
drivers of the tailgating vehicles, leading to collision \cite{freeway_oscillations}. The efficiency and solution
feasibility of rule-based systems become even worse \cite{Prakken2017-jc} in
situations with model uncertainty. Hence, we need
to take human behavior into account along with model
uncertainty for better safety guarantees.\\
We closely study a highway on-ramp merging scenario, where the goal is to enable the AV to merge with human-driven cars safely and efficiently. Our main contributions are in extending and providing a reinforcement learning (RL) pipeline with probabilistically safe constrained barrier functional safety \cite{luo2020multi}\cite{yiwei2021}\cite{cbf_theory} which can be extended to any self-driving task in a simulation domain by changing the task-based reward function. We perform experiments in CARLA (\citeauthor{Carla_Sim}) and an offline training procedure on the NGSIM dataset for a one-on-one ramp merging case. The emphasis of the paper is not on proposing a novel RL algorithm for autonomous cars, but on addressing the problem of model uncertainty with Control Barrier Functions (CBFs) (\citealt{cbf_theory}) combined with a RL pipeline to extend the research on safe RL for both offline and online environments.\\
\section{Related Work}

CBF-based methods (\citealt{cbf_theory}, \citealt{7782377}, \citealt{9196757}, \citealt{ROCBFs}) are being increasingly used
in the control application domain due to their safety guarantees
with the forward-invariant property when a solution
is feasible. In reality,a solution may not always be feasible as we deal with approximations of complex non-linear dynamics and the systems resort to hard-coded recovery control \citep{7857061}.\\
To address model uncertainty, several works \citep{9196757}\citep{nikolay2020l4dc} proposed to employ the CBF approach with noisy
system dynamics. Integrating CBF with
Model Predictive Control (MPC) is also a common planning
method. \citep{zeng2020safetycritical} and \citep{9029446} introduced
MPC-based safety-critical control. However, these works fail
to take different driving behavior styles into account. To address this problem, one possible solution is to use
CBFs with reinforcement learning-based methods, where
the agent trains for all the possible kinds of behaviors,
given enough training data and even a black-box model.
RL-CLF-CBFs proposed in \citep{choi2020reinforcement}
is used to reduce the uncertainty in plant dynamics. \citep{ma2021modelbased} implemented an RL algorithm with Constrained Policy Optimization (CPO) \citep{achiam2017constrained} and with the
added safety of control barrier functions.\\
Generalized CBFs are introduced in \citep{gcbfs} to consider higher-order
dynamics, Parametric CBFs \citep{Taylor_2020} and Exponential CBFs \citep{7524935} are popular methods to model approximate higher-order dynamics. In our work, we integrate Probabilistic CBFs \citep{pcbfs}\citep{luo2020multi}\citep{yiwei2021} to model uncertainty owing to its ability to use a simpler dynamics model and linear constraints for the optimization problem compared to other methods. \citeauthor{ma2021modelbased} is closely relevant prior work, which uses Generalized Discrete CBFs to generate safety certificates and combine with a Model Based Policy Optimization (MBPO) \citep{mbpo} framework for an intersection scenario. Although we have a similar RL pipeline to \citeauthor{ma2021modelbased}, ours is not a model-based approach and is similar to a CPO formulation. Further, we also introduced a reinforcement learning approach to CARLA and an offline NGSIM Dataset to show the generalization capabilities of our approach. 
\vspace{-5mm}
\begin{wrapfigure}[20]{r}{0.5\textwidth}
    \centering
    \includegraphics[width = 0.5\linewidth]{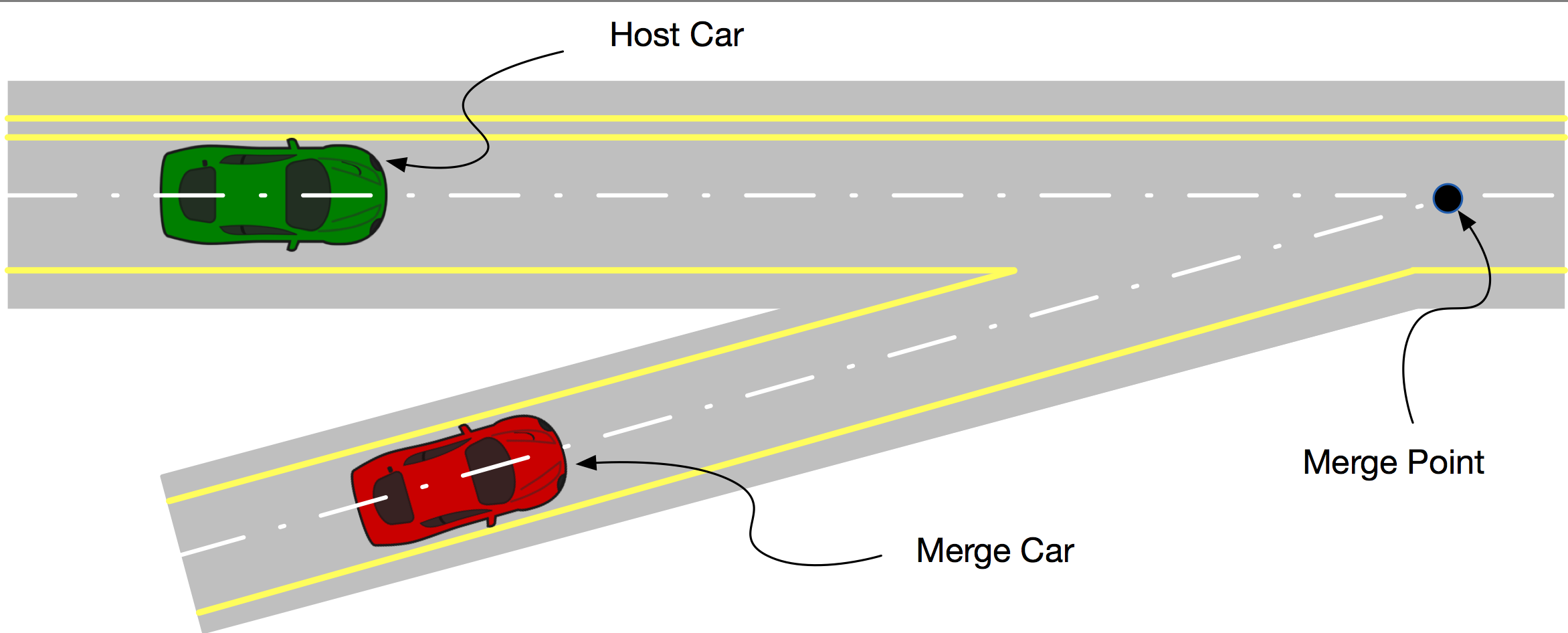}
    \includegraphics[width = 0.6 \linewidth]{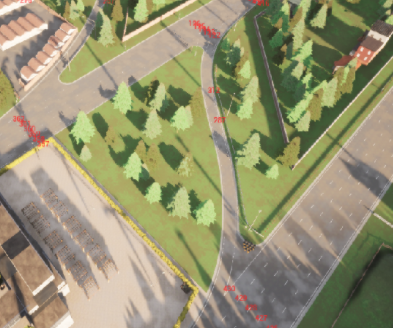}
    \caption{
    \label{scenario}
    Ramp merging scenario \citep{Dong2017:PGM}. The ego vehicle (green) is the host vehicle; the merging vehicle (red) is the ego vehicle, running on the ramp. The figure below shows a two-dimensional version of ramp-merging that we will be using. The merge is along a curve rather than a straight line.} 
\end{wrapfigure}

\section{Problem Formulation}
\subsection{Setting up the problem and Dynamics}
Ramp merging has previously been studied in one dimension. We have extended the approach to two dimensions to include a wider range of ramp-merging scenarios and easy generalization to other scenarios. The goal is to control the ego vehicle on the ramp to merge safely with the human-driven vehicles (host vehicles) already on the highway. The system dynamics of a vehicle can be described by double integrators as follows, since acceleration plays a key role in the safety considerations. 

\begin{equation} \footnotesize
\begin{split}
    \dot{X} &=\begin{bmatrix}
    \dot{x}\\
    \dot{y}\\
    \dot{v_{x}}\\
    \dot{v_{y}}
    \end{bmatrix}
    =\begin{bmatrix}
    0_{2\times2}\; I_{2\times2}\\
    0_{2\times 2} \;0_{2\times 2}
    \end{bmatrix}
    \begin{bmatrix}
    x \\
    y \\
    v_{x}\\
    v_{y}
    \end{bmatrix}
    + \begin{bmatrix}
    0_{2\times2}\; I_{2\times2}\\
    I_{2\times2}\; 0_{2\times2}
    \end{bmatrix}\begin{bmatrix}
    u_{x}\\
    u_{y}\\
    \epsilon_{x}\\
    \epsilon_{y}
    \end{bmatrix} \\
\end{split}
\label{dynamics}
\end{equation}
where $x\in\mathcal{X}\subset\mathbb{R},y\in\mathcal{Y}\subset\mathbb{R},v\in \mathbb{R}^2$ are the position and linear velocity of each car respectively and $u\in\mathbb{R}^2$ represents the acceleration control input. 
$\epsilon \sim \mathcal{N}(\hat{\epsilon}, \Sigma)$ is a random Gaussian variable with known mean $\hat{\epsilon}\in\mathbb{R}^2$ and variance $\Sigma\in\mathbb{R}^{2\times2}$, representing the uncertainty in each vehicle's motion.\
\vspace{-2mm}
\subsection{Probabilistic Control Barrier Functions}

In \citep{yiwei2021}, an optimization framework with probabilistically safe CBFs was proposed as an alternative to deterministic CBFs with perfect model information to address motion uncertainty, specifically for ramp merging scenarios in one dimension. 
Consider the admissible space equation for linear CBF with parameter $\alpha$ of one dimension, x: $\dot{h}(x,u) + \alpha h(x)\geq 0$ with $h = (x_e - x_m)^2 - R_{safe}^2$. By substituting Eq. \ref{dynamics} we have:
{\footnotesize \begin{equation}
\dot{h}^s_{em}(x,u) +\alpha h^s_{em}(x) \geq 0 ; \implies 2\Delta x_{em}^T\Delta \epsilon_{em} \geq -2 \Delta x_{em}^T(\Delta v_{em}+u_e\Delta t) - \alpha h^s_{em}(x) 
 \label{eq:Q}
\end{equation}}\!\!
where $\Delta x_{em} = x_e-x_m, \Delta v_{em}=v_e-v_m, \Delta \epsilon_{em}=\epsilon_e-\epsilon_m\sim \mathcal{N}(\Delta \hat{\epsilon}_{em}, \Delta \Sigma_{em})$ for ego vehicle $e$ and each merging vehicle $m$, $\Phi^{-1}(\cdot)$ is the inverse cumulative distribution function (CDF) of the standard zero-mean Gaussian distribution with unit variance.
We reorganize $\Pr\Big(\dot{h}^s_{em}(x, u) +\alpha h^s_{em}(x)\geq 0\Big) \geq \eta$ into the form of $\Pr(a^T c \leq b) \geq \eta \Leftrightarrow b-\Bar{a}^Tc\geq\Phi^{-1}(\eta)||\Sigma^{1/2}c||^2$ from \citeauthor{blackmore2011chance} with $a=\Delta \epsilon_{em}, c=-2\Delta x_{em}, b =2\Delta x_{em}^T (\Delta v_{em}+u_e\Delta t) + \alpha h^s_{em}(x)$ and eventually get a constraint:
$\footnotesize
        A_{em}u_x \leq b_{em},\quad A_{em}\in\mathbb{R}^{1\times 2}, b_{em}\in\mathbb{R}
$
with
{\footnotesize\begin{equation}
         A_{em} = -2\Delta x_{em}^T\Delta t;\text{ }
         b_{em} = 2\Delta x_{em}^T (\Delta v_{em}+\Delta \hat{\epsilon}_{em})
        +\alpha h^s_{em}(x) 
        -\Phi^{-1}(\eta)\sqrt{\Delta x_{em}^T \Delta \Sigma_{em}  \Delta x_{em}}
    \label{ab}
\end{equation}}
where $\Delta t$ is a time unit. We derived linear control constraints for pairwise chance-constrained safety between ego vehicle and each merging vehicle $m$. The equations in (3) can be extended to the y-dimension as well for 2-dimensional decoupled CBFs. For a more detailed derivation, we refer the reader to \citeauthor{yiwei2021} \qed
\vspace{-3mm}
\subsection{Reinforcement Learning with Probabilistically Safe CBFs}
For better safety guarantees, we use a reinforcement learning policy to compute high-level desired nominal velocity and choose a CBF control as output. However, it is computationally expensive to solve two different optimization problems while interacting with a simulator. Our constrained RL framework solves it with a single constrained optimization with linear probabilistic safety certificates.
Using the optimization framework defined in \citeauthor{achiam2017constrained} for constrained RL problems, we extend the constraints obtained in both dimensions from 3.2 for the total return objective, $\mathcal{G}$, and set of safe states, $\mathcal{C}$, as:
\begin{wrapfigure}[11]{r}{0.4\textwidth}
    \centering
    \vspace{-1.5em}
    \includegraphics[width=1\linewidth]{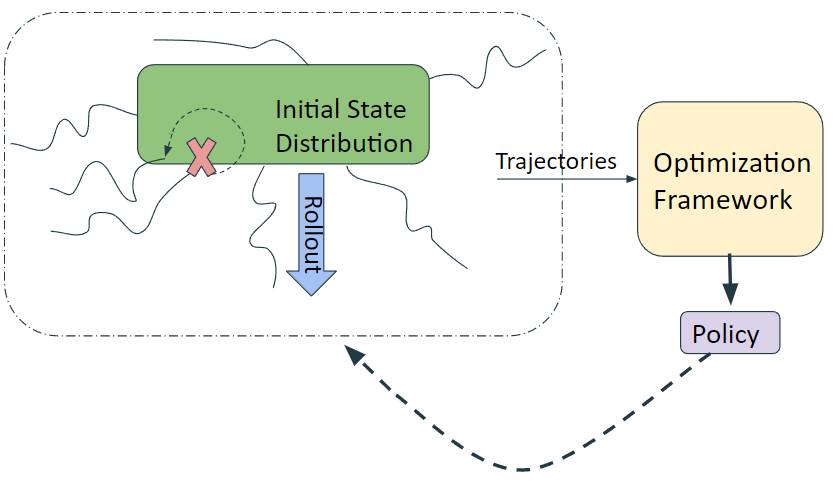}
    \vspace{-1.5em} 
    \caption{\footnotesize \textbf{Reinforcement Learning Pipeline}: We initialize rollouts from random initial states and run an optimization on the set of collected trajectories to train a policy for each epoch. }
    \label{fig:rlpipeline}
    \vspace{-1em}
\end{wrapfigure}
\vspace{-5mm}
\begin{equation}\footnotesize
\begin{split}
      &\min_{\theta} L_{a} = {\mathbb{E}_{a_{\sim} \pi,s_{\sim}C} }[\mathcal{G}] \\
    s.t \quad & {U}_{min} \leq  u_e \leq {U}_{max} \\
     & J_{c_{x}} = \mathbb{E_{u_{e} \sim \pi}}[A_{em_{x}} u_{e_{x}}] \leq b_{em_{x}}\\
     & J_{c_{y}} = \mathbb{E_{u_{e} \sim \pi}}[A_{em_{y}} u_{e_{y}}] \leq b_{em_{y}}\\
     &D_{p}(\theta,\theta_{k}) = \frac{1}{2} \Delta\theta^{T}H\Delta\theta \leq \delta
\end{split}
\label{rl_obj}
\end{equation}
Using RL objective in Eq. 4, we use the approximate actor-critic constrained policy gradient formulation in \citep{ma2021modelbased} to solve this optimization.
The critic objective is 
$L(\phi) = E_{s_{t}\sim\mathcal{C}}[\frac{1}{2}(\mathcal{G} - V(s_{t},\phi))^{2}]$, where $\phi$ is the parameters of the critic network and $V(s_t,\phi)$ is the critic's state-dependent value function.
The actor objective (Eq.4) is optimized by transforming into a dual problem and applying KKT (\citeauthor{boyd_vandenberghe_2004}) conditions. The analytical optimal solution is  $\Delta \theta^{*} = \frac{H^{-1}(g-C\nu^{*})}{\lambda^{*}}$, where $\theta$ is the actor parameters, $\lambda^*, \nu^*$ are dual variables, H is the hessian from the trust region condition, $g =\frac {dL_a}{\partial \theta}/ ||\frac{dL_a}{\partial \theta}||^2$, and $C=\frac {dJ}{\partial \theta}/ ||\frac{dJ}{\partial \theta}||^2$, combining both $J_x$ and $J_y$. If there is no feasible solution, an iterative line search method is applied using $\theta_{k+1} = \theta_k + -\sqrt{\frac{2*\delta}{C^\mathbb{T}H^{-1}C}}H^{-1}C$. A more detailed derivation can be found in the appendix (B).
From the RL pipeline in Fig. 2, we use multiple rollouts sampled from various initial states, similar to \citet{mbpo}, to capture the uncertainty in the environment and to generalize to a larger section of the state space. Algorithm 1 gives the pseudocode for our method, \textit{Safety Assured Policy Optimization for Ramp Merging} \textbf{(SAPO-RM)}.\\
\vspace{-15mm}
\begin{wrapfigure}[11]{r}{0.4\textwidth}
\vspace{-20mm}
\begin{minipage}{0.4\textwidth}
    \centering
    \begin{algorithm}[H]\footnotesize
    \caption{SAPO-RM}
    
    \begin{algorithmic}
    \REQUIRE $\Delta x_{em}, \Delta v_{em}, \Delta t, R_{safe}, {\alpha}$
    \ENSURE $\theta , \phi$
    \REPEAT
    \STATE Sample trajectories $\mathcal{D} = \mathcal{T} \sim \pi(\theta_{k})$ 
     \STATE Compute Constraints in the objective
    \STATE Estimate $\phi$ from $\mathcal{D}$
    \STATE Calculate g,C,H
    \IF {feasible}
    \STATE $\Delta \theta^{*} = \frac{H^{-1}(g+C\nu^{*})}{\lambda^{*}}$
    \ELSE
    \STATE $\theta_{k+1} = \theta_k + -\sqrt{\frac{2*\delta}{C^\mathbb{T}H^{-1}C}}H^{-1}C$
    \ENDIF
    \UNTIL convergence
    \end{algorithmic}
    \end{algorithm}
\end{minipage}
\end{wrapfigure}
\\
\vspace{-2mm}
\section{Experiments}
\vspace{-1mm}
\subsection{CARLA}
\vspace{-1mm}
\subsubsection{Setting up the Simulation Environment}
\vspace{-1mm}
We use CARLA (\citeauthor{Carla_Sim}) v.9.12 to modify the existing Town-06 environment to a ramp merging scenario. The scenario considered in Fig. 1 has a curved ramp. We use a PID Controller to track the steering control of the Ego Vehicle (On-Ramp) between waypoints generated by the default Global Planner to the specified goal position. The control for the Host (Off-Ramp) Vehicle uses both steering and longitudinal PID control. Further, the control of the host vehicle is not reactive. This assumption deals with a conservative case and satisfying this also maintains safety for the case where host control is reactive to the ego vehicle. We optimize the safe reinforcement learning objective (Eq.5) to learn a policy to control the ego-vehicle throttle position.
\subsubsection{Online Experiments on CARLA}
We implemented SAPO-RM in the Simulator set-up in 4.1.1. The RL pipeline introduced in Fig. 2 is followed, where the host vehicle is initialized at a specified initial position with a random offset. If the host vehicle reaches the destination before the ego vehicle, the host vehicle is re-initialized with random offset. The simulation is terminated only when the ego vehicle reaches the destination.The safety distance for the CBFs is 8m. Fig. 3 below shows the results for SAPO-RM for a case where the target host velocity is 40 kph. Training parameters and results for other cases are discussed in the appendix (~\ref{section:Carla_appendix}).
\vspace{-2mm}
\begin{figure}[H]
    \centering
    \includegraphics[width = 0.3 \linewidth]{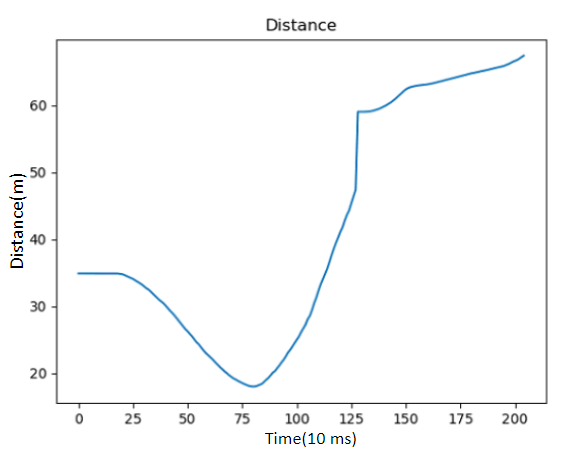}
    \includegraphics[width = 0.3 \linewidth]{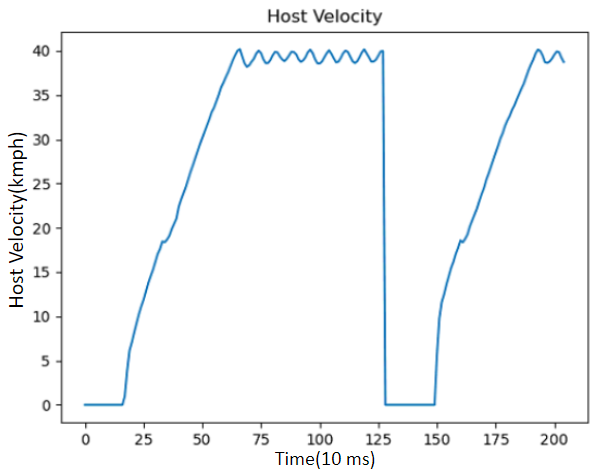}
    \vspace{1mm}
    \includegraphics[width = 0.3 \linewidth]{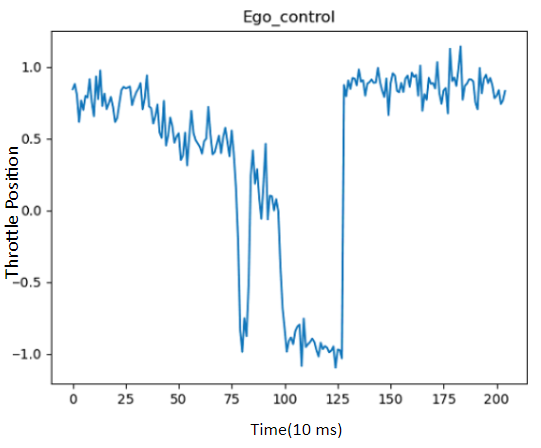}
    \caption{\textbf{Results for Online SAPO-RM:} The host velocity is 40 kph, as seen in the center image. The image on the left depicts the distance between the ego and host vehicle.The image on the right is the output of the policy for the evaluated case. From the plots of the ego control (Right) we observe that when distance is close to $R_{safe}$, the throttle position goes to full-braking, forcing the ego vehicle to maintain the safety condition. Whenever the distance again increases, the ego control goes towards full throttle to optimize performance.}
    \label{fig:my_label}
\end{figure}
\vspace{-8mm}
\subsection{NGSIM}
\subsubsection{Data Extraction}
An offline dataset for US I-80 highway ramp-merging was extracted from the NGSIM Database. We considered merging vehicles that go from on-ramp to an auxiliary ramp. To ease the complexity, the dataset was simplified into a one-on-one merging. The trajectories are first classified based on the condition: distance between host and the ego vehicle is less than 12m in the auxiliary lane. This condition leads to degenerate values as instances of freeway oscillations are observed in real data (\citeauthor{freeway_oscillations}). We constrained by sorting the data of the ego vehicle with respect to the frame number and considered the first case where the merging condition is satisfied. Further dataset extraction details are in the appendix (~\ref{section:ngsim_appendix}).
\vspace{-2mm}
\subsubsection{Implementing Offline SAPO-RM}
The offline dataset obtained in section 4.2.1 is then used to train the offline SAPO-RM policy. Similar to the online case, a rollout is initialized at a random state in the dataset. For each step in a rollout, only the parameters of the ego vehicle are updated according to the dynamics equation (Eq. 1). The vehicle heading direction ($\psi$) is used as a tracking parameter for the ego vehicle. We used a rollout length of 4 and 5000 trajectories for each epoch of training the offline RL policy for 100 epochs. In Fig. 4, we observe that the black trajectory (Ego policy) deviates from the tracking control to maintain the safe distance.
\vspace{-5mm}
\begin{wrapfigure}[16]{r}{0.4\textwidth}
    \centering
    \includegraphics[width = \linewidth]{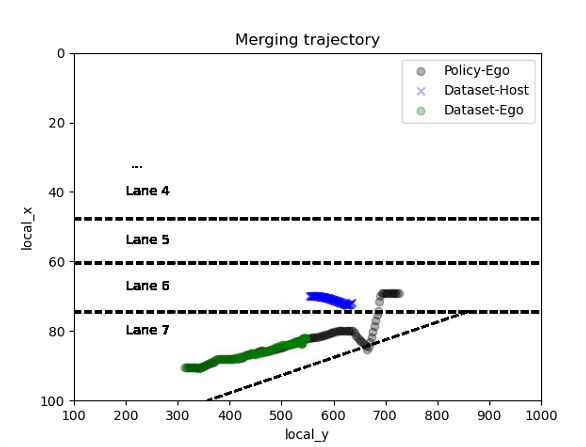}
    \vspace{-3mm}
    \caption{The black trajectory is generated by a policy, the blue one is the real host vehicle trajectory and the green one is the real ego vehicle trajectory.}
    \label{fig:my_label}
\end{wrapfigure}
\vspace{-3mm}
\section{Conclusion and Future Work}
\vspace{-2mm}
We can clearly see from Figs. 3 and 4 that safety conditions are maintained by ego-vehicles while performing ramp-merging.The algorithm we used here is not just a ramp-merging algorithm but is a safe-reinforcement learning algorithm applied to ramp-merging. By relieving some of the assumptions made on the dynamics, the safe-RL algorithm can be extended to other scenarios of lane-changing and intersections. We observe that trajectories maintain the safe distance and are efficient with SAPO-RM. Possible future extensions are extending to situations with more merging vehicles, introducing complicated dynamics models \citep{ROCBFs}, \citep{vd_model} for a better approximation and to alleviate modeling uncertainty, understanding the effect of $\alpha$ on SAPO-RM behavior, and implementing prior methods for the offline approach and other implementations to compare other methods' ability to address modeling uncertainty.


\bibliography{paper}
\bibliographystyle{icml2022}

\newpage
\appendix
\onecolumn
\section{Arriving at Probabilistic Safe Constraints}
\label{section:Probabilistic Safe Constraints}
The chance-constrained optimization framework introduced in \cite{yiwei2021} to handle stochastic world dynamics can be extended into two dimensions, x and y, explicitly in our implementation to accommodate uncertainty in the model information. The equation below shows the CBF-QP with a probabilistic chance constraint in one dimension:

\begin{equation}\footnotesize
\begin{split}
      &\min_{u_e\in\mathcal{U}_e} || u_e-\Bar{u}||^2 \\
    s.t \quad & {U}_{min} \leq  u_e \leq {U}_{max} \\
     &\Pr\Big(\dot{h}^s_{em}(x, u) +\alpha h^s_{em}(x)\geq 0\Big) \geq \eta,\qquad \forall m
\end{split}
\label{qp}
\end{equation}
where $\Bar{u}$ is the nominal expected acceleration for the ego vehicle to follow, and $U_{max}$ and $U_{min}$ are the ego vehicle's maximum and minimum allowed acceleration. 
 $\text{Pr}(\cdot)$ denotes the probability of a condition to be true. $\Pr\Big(\dot{h}^s_{em}(x, u) +\alpha h^s_{em}(x)\geq 0\Big) \geq \eta\implies \Pr\Big(h^s_{em}(x)\geq 0\Big) \geq \eta$ given the forward invariance set theory in a deterministic setting, $\dot{h}(x, u) +\alpha h(x)\geq 0\implies h(x)\geq 0$ as proved in \cite{ames2019control} through a chance constraint.
 
  From \cite{blackmore2011chance}, a general chance constraint problem can be transformed to a deterministic constraint as 
\begin{equation}\label{eq:chance}\footnotesize
\begin{split}
        &\Pr(a^T c \leq b) = \Phi(\frac{b-\Bar{a}^T c}{\sqrt{c^T\Sigma c}})\\
        \implies &\Pr(a^T c \leq b) \geq \eta \Leftrightarrow b-\Bar{a}^Tc\geq\Phi^{-1}(\eta)||\Sigma^{1/2}c||^2
\end{split}
\end{equation}

\section{Constrained Policy Gradient Approximation}
\label{section:Constrained_RL}
The Reinforcement Learning objective of the actor in Eq. 4 can be transformed into a dual objective as:
\begin{equation}\footnotesize
\begin{split}
    &\min_{\Delta \theta} g^{\mathbb{T}}\Delta \theta\\
    &s.t.\text{  } z + C^{T}\Delta\theta \leq 0\\
    & \Tilde{D_{p}} \sim \frac{1}{2} \Delta^{\mathbb{T}}H\Delta\theta \leq \delta\\
    \end{split}
\end{equation}

where $g =\frac {dL_a}{\partial \theta}/ ||\frac{dL_a}{\partial \theta}||^2$,$C=\frac {dJ}{\partial \theta}/ ||\frac{dJ}{\partial \theta}||^2$, $z = J_c - b_m$. Using a Lagrange multiplier, the Lagrangian function then is:
$$
\mathbf{L(\theta,\nu)} = g^{\mathbb{T}}\Delta\theta + \lambda(\frac{1}{2}\Delta\theta^{\mathbb{T}}H\Delta\theta - \delta) + \nu(z + C^{\mathbb{T}}\Delta\theta)
$$
$\lambda ,\nu$ are dual variables. Using KKT Conditions \citep{boyd_vandenberghe_2004},we get the equations:
\begin{align*}\footnotesize
    &\frac{\partial \mathbf{L}}{\partial \Delta\theta} = g + \lambda H\Delta\theta + \nu C \\
    &\lambda(\frac{1}{2}\Delta\theta^{\mathbb{T}}H\Delta\theta - \delta) = 0\\
    & \nu(z + C^{\mathbb{T}}\Delta\theta) = 0\\
    & \lambda,\nu \geq 0\\
    & (\frac{1}{2}\Delta\theta^{\mathbb{T}}H \Delta \theta - \delta) \leq 0\\
    & (z + C^{\mathbb{T}}\Delta\theta) \leq 0
\end{align*}
\label{rl_obj_1}
solving which, an optimal update direction can be obtained. If there is a feasible solution, the optimal update direction is $$\Delta \theta^{*} = \frac{H^{-1}(g-C\nu^{*})}{\lambda^{*}} $$
where $\nu^*,\lambda^*$ are optimal dual variables. For the retrieval update, we ignore the objective function and take the gradient descent with respect to the constraints to force the policy back to the safety region. Similar to \citep{achiam2017constrained}, $\theta_{k+1} = \theta_k + -\sqrt{\frac{2*\delta}{C^\mathbb{T}H^{-1}C}}H^{-1}C$ is the retrieval policy.\\
The check for feasibility can be done by solving the optimization problem, proposed in \citep{Duan_2022}:
\begin{equation}\footnotesize
\begin{split}
    &\min_{\Delta \theta} \Delta\theta^T H \Delta\theta\\
    &s.t.\text{  } z + C^{T}\Delta\theta \leq 0\\
    \end{split}
\end{equation}

If the optimal value of the optimization problem above is $\delta_{min}$, the feasible solution set is empty if $\delta_{min} \geq \delta$ and contains a solution otherwise. This problem is optimized efficiently using the Lagrangian Dual Problem:
$$
\max_{\nu \geq 0}  \frac{-\nu^TC^TH^{-1}C\nu}{2} + \nu^Tz
$$
A feasibility check can be done by comparing $\delta_{min}$ with $\delta$.
\section{Running experiments with CARLA Environments}
\label{section:Carla_appendix}
\subsection{Training Parameters and Environment Setup}
The state dimension is 37. Since we have to reproduce the entire state of an initial state distribution, we save the positions, velocities, angular velocities,  throttle and steering positions in three dimensions of both the ego and merging vehicle.The action dimension is 1, as we are only controlling the throttle position. We use 100 trajectories with a rollout length of 20. The actor policy has 2 MLP layers with a size of 256, and the critic is also MLP of size 256. The parameter $\alpha$ is set to 0.75, the uncertainty parameter $\eta$ is 0.99, and $\Delta t$ is 0.01. The reward function we use is the negative weighted sum of error from the current position to the merging position chosen on the Town-06 environment map and the error of the velocity to the desired velocity. We used a desired velocity of 35 kmph for the ego-vehicle. There are no bounds on the throttle position, other than the specified range in Carla from [-1,1].
\subsection{Variation with Host Velocity}
We saw the results for the host velocity of 40 kph in section 4.1.2. In the figure below, we can see the results with the cases where the host velocities are 30 kph and 20 kph. From the Figures 5 \& 6 below it is evident that the minimum distance of 8m is respected with a range of initial conditions that we train our policy on. However, once we step outside the training distribution, we cannot guarantee that this will generalize.
\begin{figure}[H]
    \centering
    \includegraphics[width = 0.3 \linewidth]{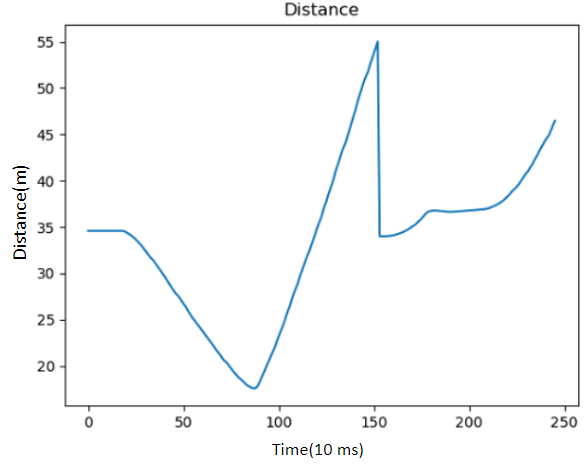}
    \includegraphics[width = 0.3 \linewidth]{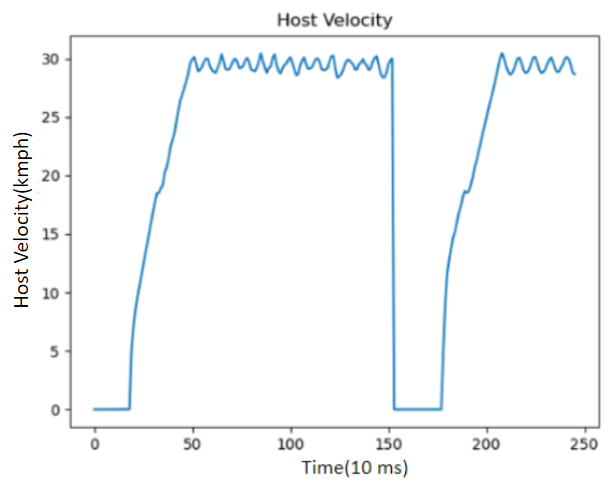}
    \vspace{1mm}
    \includegraphics[width = 0.3 \linewidth]{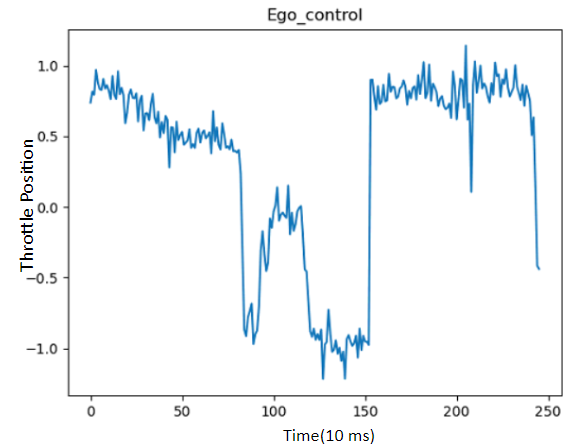}
    \caption{\textbf{Host velocity - 30 kmph}}
    \label{fig:my_label}
\end{figure}
\begin{figure}[H]
    \centering
    \includegraphics[width = 0.3 \linewidth]{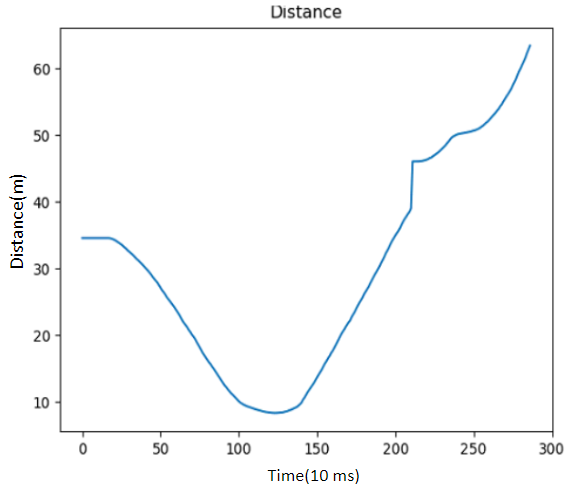}
    \includegraphics[width = 0.3 \linewidth]{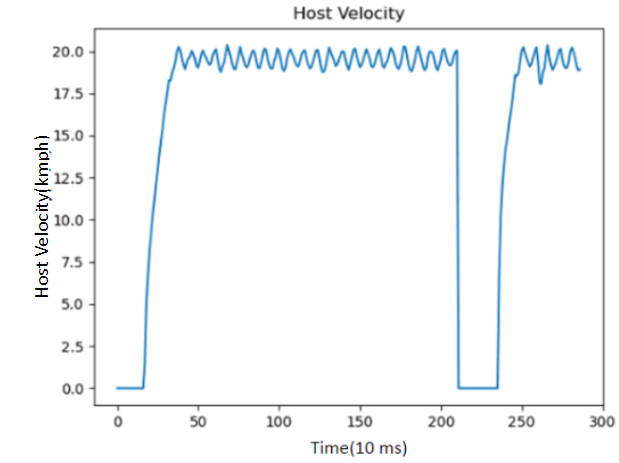}
    \vspace{1mm}
    \includegraphics[width = 0.3 \linewidth]{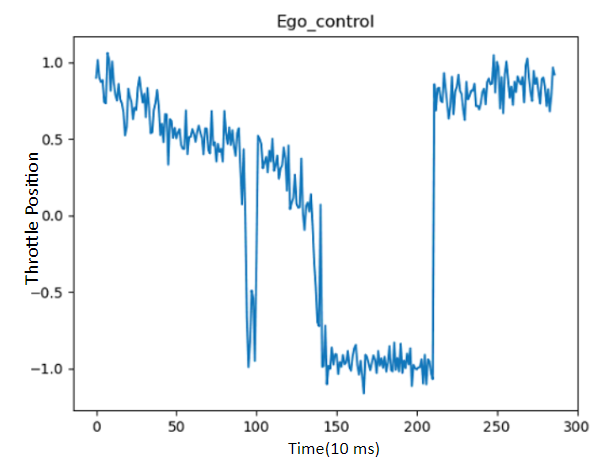}
    \caption{\textbf{Host velocity - 20 kmph}}
    \label{fig:my_label}
\end{figure}
\subsection{Variation with $\alpha$ parameter of the CBF}

In Figure 7, we repeat the experiment of the ramp merging case with Host velocity 20 kph, as observed in Figure 6. With a higher value of alpha, ideally we should observe more aggressive behavior, which is also the case from the plots in Fig. 7. One main difference between the two versions is that the merging agent waits before the merge for $\alpha = 0.75$ and merges behind the host vehicle. For the case with higher $\alpha$, the merging vehicle accelerates from the beginning and would merge the host vehicle in front of it.
\begin{figure}[H]
    \centering
    \includegraphics[width = 0.3 \linewidth]{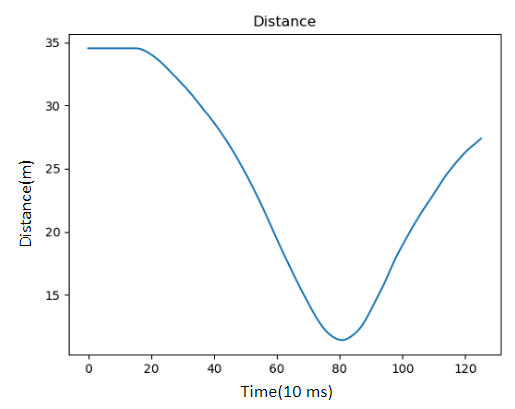}
    \includegraphics[width = 0.3 \linewidth]{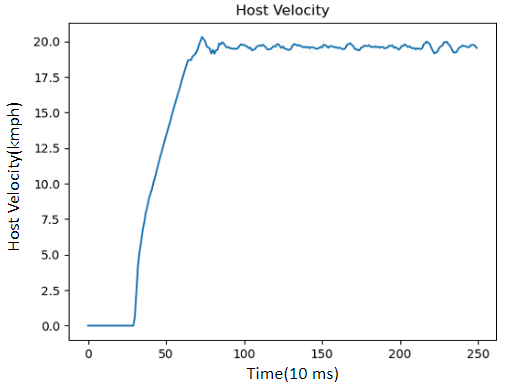}
    \vspace{1mm}
    \includegraphics[width = 0.3 \linewidth]{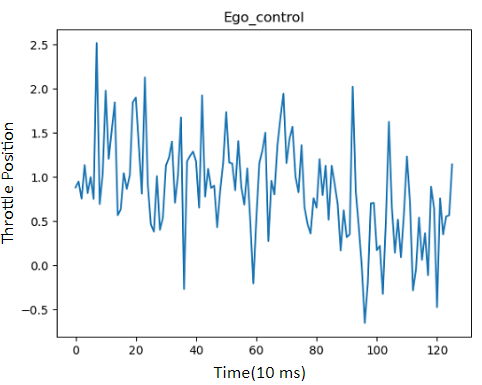}
    \caption{\textbf{Host velocity - 20 kmph and $\alpha$ = 15}}
    \label{fig:my_label}
\end{figure}
\pagebreak
\section{NGSIM Dataset Extraction}
\label{section:ngsim_appendix}
Of the possible 11,850,526 data points in the NGSIM Dataset we have extracted a train, validation and test split of the merging dataset of about 100,000 data points of the merging trajectories for an one-on-one situation. The figure below shows the trajectories of the merging condition we have applied on the US I-80 highway dataset. We filter the ego vehicles that start in lane 7. Of the possible combinations to go to lane 5 and lane 6, we only consider the case where vehicles merge into lane 6. For the host vehicles, we consider the vehicles that are within a specified distance of 12m, when the ego vehicle is merging into lane 6. We also sort the data in terms of frame numbers of ego vehicles for time-based trajectories. The trajectories generated can be seen in Figure 7. The network parameters of actor and critic, and the parameters of $\alpha, \eta \text{ and} \Delta t$ are the same as that for the online case. The reward function in this case is negative tracking error between the current trajectory and the dataset trajectory for ego-vehicle.
More results from different initial positions are shown in Figure 8. We will report more results on the average increase in distance and time performance gain on the dataset in our future work.
\begin{figure}[t]
    \centering
    \includegraphics[width = 0.6 \linewidth]{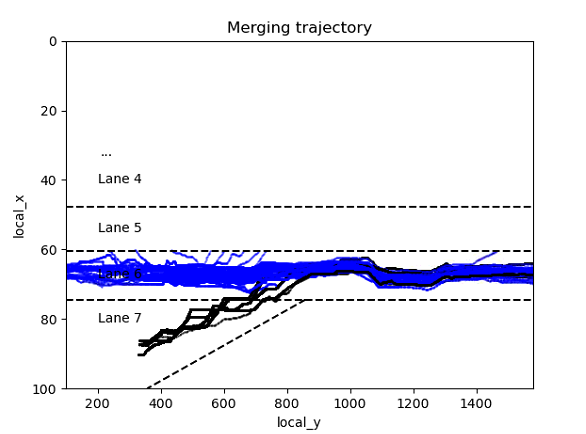}
    \caption{\textbf{Merging Dataset:} It depicts all the trajectories extracted for the I-80 dataset for one-on-one ramp merging.Trajectories in blue are that of the host vehicle and in black are of the ego vehicle.}
    \label{fig:ngsim}
\end{figure}
\vspace{-10mm}
\begin{figure}[H]
    \centering
    \includegraphics[width = 0.4 \linewidth]{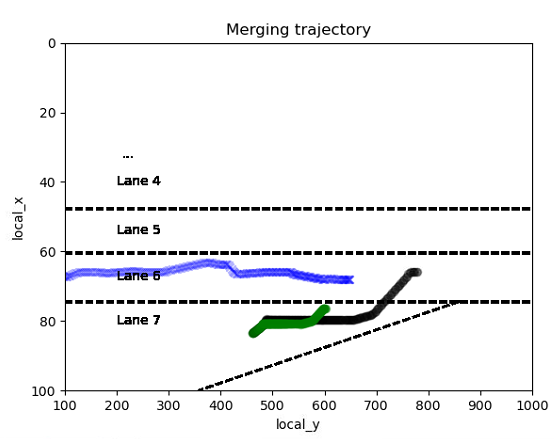}
    \includegraphics[width = 0.4 \linewidth]{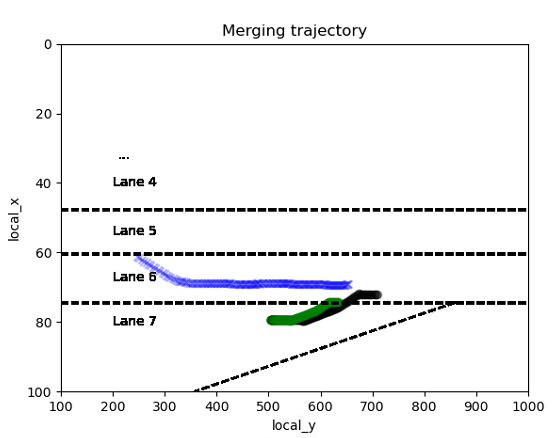}
    \caption{\textbf{Offline Sapo RM:} The results trajectories shown here compare the green and black trajectories, which are the orignal trajectory in the dataset and the RL policy trajectory respectively.}
    \label{fig:ngsim}
\end{figure}

\begin{figure}[H]
    \centering
    \includegraphics[width = 0.4 \linewidth]{ngsim_1.png}
    \includegraphics[width = 0.4 \linewidth]{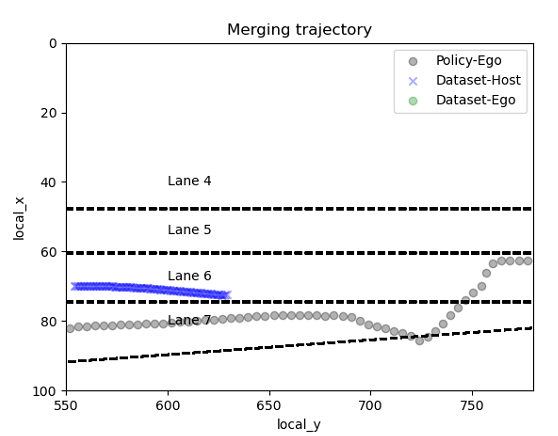}
    \caption{\textbf{Observing Trajectory:} On the image to the left we observe the entire trajectory of Fig.4 and on the figure on the right we looked at a zoomed in version to observe the same trajectory in more detail.As the figure on the left has a different scale, it appears that the trajectories are very sharp but with a detailed trajectory on the right, the change in the heading direction of the car is not very sharp but rather an illusion of scale.}
    \label{fig:ngsim}
\end{figure}

\end{document}